\documentclass[runningheads]{llncs}

\usepackage{subcaption}
\usepackage{paralist}
\usepackage{makecell,multirow,multicol,booktabs,amsmath,amsfonts}
\usepackage{graphicx}
\usepackage{cite}
\usepackage{xcolor}
\usepackage[linesnumbered,ruled,vlined]{algorithm2e}
\usepackage{eso-pic}

\AddToShipoutPictureBG{
  \AtPageUpperLeft{
    \raisebox{-2\baselineskip}{\makebox[\paperwidth]{\begin{minipage}{21cm}\centering
      Paper accepted at the 29th European Symposium on Research in Computer Security (ESORICS)
    \end{minipage}}}
  }
}

\graphicspath{{./img/}}

\DeclareMathOperator*{\argmin}{arg\,min}
\SetKwInput{KwInput}{Input}             
\SetKwInput{KwOutput}{Output}

\begin{document}

\title{Have You Poisoned My Data? Defending Neural Networks against Data Poisoning
\thanks{Partially funded by the Technology Innovation Institute (UAE) under the project ``Prevention of Adversarial Attacks on Machine Learning Models'', the PON program of the Italian MUR under the project ``Application of Machine Learning to improve olive yield and reduce climate change impact'', and project SERICS (PE00000014) under the NRRP MUR program funded by the EU-NextGenerationEU.}}

\author{Fabio De Gaspari\inst{1}\orcidID{0000-0001-9718-1044} \and
Dorjan Hitaj\inst{1}\orcidID{0000-0001-5686-3831} \and
Luigi V. Mancini\inst{1}\orcidID{0000-0003-4859-2191}}
\authorrunning{F. De Gaspari et al.}

\institute{Dipartimento di Informatica, Sapienza University of Rome, Rome, Italy
\email{\{surname\}@di.uniroma1.it}}

\maketitle

\begin{abstract}
The unprecedented availability of training data fueled the rapid development of powerful neural networks in recent years. However, the need for such large amounts of data leads to potential threats such as poisoning attacks: adversarial manipulations of the training data aimed at compromising the learned model to achieve a given adversarial goal. 

This paper investigates defenses against clean-label poisoning attacks and proposes a novel approach to detect and filter poisoned datapoints in the transfer learning setting. We define a new characteristic vector representation of datapoints and show that it effectively captures the intrinsic properties of the data distribution. Through experimental analysis, we demonstrate that effective poisons can be successfully differentiated from clean points in the characteristic vector space.
We thoroughly evaluate our proposed approach and compare it to existing state-of-the-art defenses using multiple architectures, datasets, and poison budgets. Our evaluation shows that our proposal outperforms existing approaches in defense rate and final trained model performance across all experimental settings.

\end{abstract}

\section{Introduction}
The recent widespread success of machine learning rests in no small part on a large amount of public data available for training. Cutting-edge Deep Neural Networks (DNN), such as DALL-E, Imagen, LLaMA-2, and GPT4, have up to tens of billions of parameters trained by scraping as much data from the Internet as possible. The ever-growing amount of data required to train such large models makes it impractical to carefully filter and select what is included in the training set, especially in distributed learning settings~\cite{hitaj2023fedcomm}, opening the doors to training-time adversarial attacks. One such family of attacks is data poisoning. Poisoning attacks manipulate the training dataset by injecting or maliciously altering datapoints, compromising the learned model to achieve a predefined adversarial goal. The goal of poisoning attacks can be typically divided into three categories~\cite{cina2023wild}: (1) integrity violation, where the adversary aims to preserve the trained model's performance, altering its output only under specific conditions;  
(2) availability violation, where the poisoning attack is crafted to decrease the performance of the trained model on its intended task; (3) privacy violation, where data poisoning is used to force the model to leak private information about the system, its users, or dataset. 
\vspace{-0.1em}

This paper focuses on category (1) poisoning attacks, which involve compromising the integrity of the trained model to force misclassification for specific queries.
In particular, we consider a particularly dangerous family of attacks documented in recent scientific literature: \emph{triggerless clean-label poisoning attacks}. Triggerless clean-label attacks apply a constrained perturbation to a subset of the training set so that the perturbed samples reside closely to a target sample that the attacker wants to misclassify. The measure of closeness between the perturbed and target samples and the space in which their distance is measured can vary based on the specific poisoning attack. Some attacks force a collision between perturbed samples in the feature space of the model~\cite{shafahi2018poison,zhu2019transferable,aghakhani2021bullseye}, while other proposals work in the gradient space~\cite{geiping2020witches}. Regardless of the specific process used to craft the perturbation, the goal is to force the model to misclassify a given target sample without injecting any obvious triggers or altering the labels of the training data~\cite{shafahi2018poison}. We focus on triggerless clean-label poisoning attacks because their characteristics make them appealing to adversaries. 
Firstly, the adversarial perturbation applied to poisoned samples is heavily constrained, making it hard to detect with traditional approaches such as L2-norm~\cite{peri2020deep}. Secondly, unlike backdoor attacks that rely on injecting a trigger in the query sample during inference~\cite{liu2018trojaning,nguyen2020input}, triggerless clean label attacks do not require modification of the target sample at inference time. Thirdly, the attack is stealthy as there is typically minimal performance impact on a poisoned model, making it hard to detect by model performance analysis. Finally, since the label of the poisoned training samples is unaltered and the perturbation is heavily constrained, poisons appear normal and are challenging to spot even with expert human inspection.
Given the dangers of deploying a potentially poisoned model, especially in critical domains~\cite{10262059}, several defense mechanisms have been proposed in recent years~\cite{tian2022comprehensive}. However, current defenses have significant shortcomings, mainly falling into four categories: failure to generalize, failure against strong poison generation algorithms, performance degradation, and failure against large adversarial budgets. Many defenses are designed against specific poison-crafting attacks and fail to generalize to different approaches~\cite{paudice2019label,tran2018spectral,peri2020deep,chen2018detecting}. Other techniques are effective against some poisoning attacks, but fail against stronger poison-generation algorithms~\cite{koh2022stronger,shokri2020bypassing}. Some proposed defenses effectively prevent the models' poisoning but negatively impact testing and generalization performance~\cite{yang2022not,geiping2021strong}. Finally, as we show later in our evaluation, some defenses fail when the adversary is allowed a large poison budget (the portion of the training set that is poisoned) or perturbation budget (the constraint on the amount of allowed perturbation).

We address these shortcomings and propose a new defense method to sanitize the training set and filter poisoned datapoints in transfer learning settings. In transfer learning, a pre-trained network is used as a feature extractor to train another downstream network on a given task. Transfer learning allows repurposing the knowledge learned by the pre-trained network to provide more meaningful features to another network, without the need to train it from scratch.  We focus on the transfer learning setting because it is quickly becoming the most common use case in deep learning. The large number of parameters of contemporary models and the immense dataset requirements make it impractical to train models from scratch~\cite{touvron2023llama}. On the other hand, the widespread availability of large, pre-trained models keeps increasing, with companies such as Meta releasing open-source, cutting-edge neural networks to the public~\cite{llama2,codellama}. Finally, poisoning attacks are considerably more effective in a transfer learning scenario. The pre-trained extractor allows crafting more effective adversarial perturbations, which make it easier to poison the downstream network during fine-tuning~\cite{schwarzschild2021just}.
In light of these considerations, we propose a new poison sanitization approach based on the analysis of low- and high-level feature maps of the samples in the dataset. 
We hypothesize that the perturbation injected by poisoning algorithms is sufficient to meaningfully shift the distribution of poisons from clean images at different levels of representation within the network. We relate this hypothesis to a recent work on image synthesis~\cite{yin2020dreaming}, where Batch Normalization (BN) layers are used to effectively characterize the distributions of different classes in the dataset. We build on this insight and design a new characteristic vector representation to describe datapoints. We exploit this representation to detect poisons by measuring the distance between the datapoints in the dataset and a centroid pseudo-datapoint, which represents the general characteristics of each individual class. 
Effectively, we use BN layers as a proxy to describe the characteristics of low- and high-level feature maps of datapoints and distinguish samples drawn from clean and poisoned distributions in the characteristic vector space.
We carry out a thorough experimental evaluation and demonstrate that, given a robustly trained feature extractor, characteristic vectors can be used to recognize poisons effectively. We show that our approach generalizes to multiple poison-generation techniques, is robust against strong poisons, does not affect the model's performance, and is resilient against high poison perturbation budgets. We experimentally compare against recently proposed poisoning defenses and show that our approach outperforms the state-of-the-art in test accuracy and success rate. Moreover, we show that our approach can successfully separate real poisons from failed poisons: poisoned datapoints that do not affect the model's learned decision boundary. Summarizing, this paper makes the following contributions:
\vspace{-0.3em}
\begin{itemize}
    \item We propose a novel approach to effectively separate clean and poisoned samples in a training dataset. We rely on BN layers as a proxy to summarize the characteristics of low- and high-level feature maps of datapoints and build a characteristic vector representation to separate poisons from clean samples.

    \item We demonstrate that characteristic vectors are strong distinguishers for poisons. We show that our characterization allows the effective separation between real poisons and failed poisons (referring to poisoned datapoints that do not impact the model's learned decision boundary). Furthermore, we illustrate that clean datapoints are distinctly separated from real poisons within the characteristic vector space.

    \item We show that while failed poisons overlap with clean points of the same class, real poisons fall in the class manifold of the target class in the characteristic vector space, i.e., the class the attacker wants to misclassify a sample as.
    
    \item We thoroughly evaluate our approach and show that it consistently outperforms current state-of-the-art defenses in test accuracy and success rate. Through extensive experimental evaluation, we demonstrate that our approach generalizes to several poison-generation algorithms and is resilient against high poison and perturbation budgets.
    
\end{itemize}
\section{Background}
\label{sec:background}

Triggerless, Clean-label poisoning attacks (\emph{clean label attacks}, from here on) are training-time DNN attacks that manipulate the training set to alter the learned decision boundary and cause the misclassification of a predefined target sample at inference time. Triggerless clean-label attacks have four main characteristics: 
(1) their applied perturbation is constrained; (2) no trigger~\cite{liu2018trojaning} is added to the samples during training nor inference (3) they do not change the label of the poisoned samples; (4) they do not degrade the performance of the trained model. These characteristics make clean-label attacks particularly dangerous and hard to detect. Clean-label attacks randomly sample a small set of datapoints from a given class in the training set, called \emph{base class}, and apply a constrained perturbation to these samples. The perturbation is crafted so that a DNN trained on the poisoned images misclassifies a given \emph{target image} to the selected base class. For instance, a clean-label attack on a cats vs. dogs classifier will perturb a random set of cat images so that a specific dog image is classified as a cat.

Formally, clean-label poisoning can be formalized as a bilevel optimization problem. Let $f(x, \theta): \mathbb{R}^n \rightarrow \mathbb{R}^m$ be a machine learning model with inputs $x \in \mathbb{R}^n$ and parameters $\theta \in \mathbb{R}^p$. Let $\mathcal{L}$ denote a chosen loss function, $D_{train} = \{ (x_i, y_i) | 1 \leq i \leq N\}$ the training dataset, and $P \subset D_{train}$ a subset of $k = \| P \|$ poisoned datapoints of class $y^b$, called the \emph{base class}. The adversarial task is to optimize a constrained perturbation $\Delta_i$ for each datapoint in $P$ such that a given target sample $x^t \notin D_{train}$ with real label $y^t$ is classified by $f$ as the base class $y^b$:

\vspace{-1.5em}
\begin{align}
\label{eq:bilevel}
    \argmin_\Delta \;\mathcal{L}( f(x^t, \theta_\Delta), y^b)
    \hspace{4em}
    \argmin_\theta \;\frac{1}{N}\sum_{i=1}^N \mathcal{L}(f(x_i + \Delta_i, \theta), y_i)
\end{align}
\vspace{-0.5em}
\begin{equation*}
\begin{split}
    \text{s.t.}& \\
    \| \Delta_i \| &\leq \epsilon \;\; \forall x_i \in P \\
    \Delta_i &= 0 \;\; \forall x_i \in D_{train} \setminus P \\
\end{split}
\end{equation*}

where $\| \cdot \|$ is a norm function (typically, l-infinity norm) and $\theta_\Delta$ are the parameters of the model trained on the perturbed datase. The minimization in the LHS of Eq.~\ref{eq:bilevel} ensures that the trained model $f(\theta_\Delta)$ misclassified the target sample by minimizing the loss between $x^t$ and the base label $y^b$, while the RHS of Eq.~\ref{eq:bilevel} ensures the network is properly trained on its task.

\subsection{Feature Collision}
\label{sec:fc}
Feature Collision (FC) poisons~\cite{shafahi2018poison} are clean-label poisons crafted so that the poisoned base images lie close to the target image in the feature space of a target model. Formally, feature collision poisons are generated by solving the following optimization problem:

\begin{equation}
\label{eq:fc}
    x_i^p = \argmin_x \| f(x, \theta) - f(x^t, \theta) \|_2^2 + \beta \| x - x_i^b \|_2^2
\end{equation}

The original construction presented in equation~\ref{eq:fc} uses a weak constraint on the allowable perturbation through the penalization term $\beta \| x - x_i^b \|_2^2$. In practice, this constraint is not sufficient to guarantee that the generated poisons are clean-label\cite{schwarzschild2021just}, and an l-inf norm constraint is preferred: $\| x_i^p - x_i^b \|_{inf} \leq \epsilon$.

\subsection{Convex Polytope and Bullseye Polytope}
\label{sec:cp}
Convex Polytope (CP) poisons~\cite{zhu2019transferable} use a relaxed constraint for poison generation compared to FC. Rather than forcing a collision between poisons and the target image, CP poisons are crafted such that the feature representation of the target is a convex combination of the feature representations of the poisoned samples. Bullseye Polyotpe (BP) poisons~\cite{aghakhani2021bullseye} reduce the computational complexity of computing the convex polytope by fixing some coefficients of the original CP formulation, increasing robustness and generalization. Formally:

\begin{equation}
\begin{split}
    x^p = \argmin_{x_i} &\frac{1}{2m} \sum_{j=1}^m \frac{\| \phi_j(x^t) - \frac{1}{k} \sum_{i=1}^k \phi_j(x_i) \|^2}{\| \phi_j(x^t) \|}\\
    \text{s.t.} \;\; &\| x_i - x_i^b \|_{inf} \leq \epsilon \; \forall i  \in [1,k]
\end{split}
\end{equation}

Since BP is a strict improvement over CP~\cite{schwarzschild2021just}, in this work we only consider BP poisons.

\subsection{Gradient Matching}
\label{sec:gm}
Gradient Matching (GM) poisons~\cite{geiping2020witches} craft a perturbation such that the gradient of the poisons during training aligns with the gradient of the target image by minimizing their negative cosine similarity:

\begin{equation}
\begin{split}
    \argmin_{\Delta_i} &1 - \frac{\langle \nabla_\theta \mathcal{L}(f(x^t,\theta), y^b) \sum_{i=1}^k\nabla_\theta \mathcal{L}(f(x_i^b+\Delta_i, \theta), y_i^b)\rangle} 
    {\| \nabla_\theta \mathcal{L}(f(x^t,\theta), y^b) \| \cdot \| \sum_{i=1}^k \nabla_\theta \mathcal{L}(f(x_i^b+\Delta_i, \theta), y_i^b) \|}\\
    \text{s.t.} \;\; &\| \Delta_i \|_{inf} \leq \epsilon \; \forall i \in [1,k]
\end{split}
\end{equation}

The idea behind GM is that aligning the gradient of poisons and targets is sufficient to cause the learned model to misclassify a given target image.
\section{Threat Model}
\label{sec:threat_model}
We consider a standard transfer learning setting. A user (victim) has access to a model $\phi$ that is pre-trained on a task $\mathbb{A}$. The user has access to a training dataset $D_{train}$ and wants to use $\phi$ as a feature extractor to train another model $f$ on task $\mathbb{B}$ which is related to $\mathbb{A}$.
Consistently with the literature on clean-label poisoning~\cite{peri2020deep,yang2022not}, we consider an adversary with limited access to the training dataset $D_{train}$. Such an adversary is unable to insert or remove datapoints from the train set, but can alter a subset of the training datapoints $P \subset D_{train}$ by injecting them with a constrained perturbation.
This altered subset of datapoints is called the \emph{poison set}. The number of datapoints the adversary is allowed to poison is called the \emph{poison budget}, and the constraint on the amount of perturbation allowed on each datapoint is called the \emph{perturbation budget}. The poison set $P$ is altered by the attacker to be \emph{clean-label}: the perturbation injected by the adversary does not change the label that a human observer would give to the datapoint. For example, an image of a boat altered with a clean-label poison attack would still be labeled as a boat by a human observer. The goal of the adversary is to create a poisoned set $P$ using training images from a given \emph{base class} $y^b$ such that, when the DNN $f$ is trained on $\phi(x_i) \forall x_i \in D_{train}\cup P$, $f$ will misclassify a \textit{target sample} $x^t$ as the given base class chosen by the adversary: $f(\phi(x^t)) = y^b$. We consider the best possible scenario for the attacker, with full knowledge of the training data $D_{train}$, training procedure, and feature extractor $\phi$ used by the victim.

The victim has no knowledge of any details of the attack. In particular, we assume no knowledge of the target sample $x^t$ or base class $y^b$ chosen by the adversary, nor any knowledge regarding the poison budget or perturbation budget. Finally, we assume the victim has no access to any training data other than $D_{train}$ and no knowledge of any known clean datapoints in $D_{train}$.

\section{Our Approach}
Several existing approaches rely on the analysis of the feature-space representation of datapoints at the last layer of the network to detect poisons. The rationale behind these approaches is that the feature space representation of the poisoned points diverges from that of clean points, and this divergence can be detected with different means (e.g, KNN in Peri et al.~\cite{peri2020deep}, Spectral Signatures in Tran et al.~\cite{tran2018spectral}). While this assumption generally holds true for some poison generation algorithms that explicitly promote this objective, such as FC~\cite{shafahi2018poison} and CP/BP~\cite{zhu2019transferable,aghakhani2021bullseye}, it does not always hold for other techniques such as GM~\cite{geiping2020witches}. Moreover, since these techniques are designed to detect feature space deviations from the majority distribution, they are effective only when adversaries are allowed low poison budgets.

The key observation behind our approach is that, in order to minimize Eq.~\ref{eq:bilevel}, poison optimization algorithms are incentivized to push low- and high-level feature maps of the poisons toward the target class across all layers of the DNN. Building on observations in previous works on image synthesis~\cite{yin2020dreaming}, we use the information encoded in the Batch Normalization (BN) layers to characterize the feature distribution of the classes in the dataset at different depths of the network. Based on this characterization, we build a \emph{characteristic vector} for each datapoint in $D_{train}$ and measure its distance to the characteristic vector of a centroid pseudo-datapoint computed for every class. Finally, we detect mismatches between such distance and the class label assigned to the sample. The characteristic vector is a vector encoding BN statistics for a datapoint (or group of datapoints) at different levels of representation (i.e., depths) within the network. Effectively, our approach does not measure the deviation of poisons from the base class (i.e., the class of the datapoints used to generate the poisons), which can be easily influenced by large perturbation budgets or different poison generation techniques. Rather, we measure the \emph{convergence} of the poisons \emph{toward} the target class, which is required for the attack to be successful (see Section~\ref{sec:eval_distance}). Furthermore, we do so using features that are robust and that any poison generation technique necessarily modifies to cause misclassification. As a result, our poison detection approach is resilient to large poison and perturbation budgets, and generalizes across poison generation algorithms that use different optimization goals, as demonstrated in our experimental evaluation. In the following sections, we present a formal description and discuss the implementation details of our poison detection approach.

\subsection{Formal Description of the Approach}
\label{sec:formal_description}
Let $\phi$ be a pre-trained feature extractor with $l$ layers, $D_{train} = \{ (x_j,y_j) \mid j<N\}$ our training dataset, $Y$ the classes of the datapoints, and $P \subset D_{train}$ a subset of $k = ||P||$ poisoned datapoints. Let $L^{bn}_i \;\forall i<l$ be the i-th batch normalization layer of $\phi$ and $\mu_i(X),\sigma_i(X)^2$ the channel-wise mean and variance of $L^{bn}_i$ computed over a given set of datapoints $X$. We first compute the \emph{centroid characteristic vector} of the distribution for each class in the training set 

\begin{equation}
    \mathcal{C}_y = \{(\mu_i(X_y), \sigma_i(X_y)^2) \mid \forall i < l\} \;\forall y \in Y
\end{equation}

where $X_y$ is the set of all the datapoints in $D_{train}$ with label $y$. For the poisoned class, this includes the poisoned samples $P$ together with the clean samples. The centroid characteristic vector provides a summary of the characteristic features of each class in the dataset at different levels of representation within the pre-trained network $\phi$. For each datapoint in the training set, we compute their characteristic vector $\mathcal{X}_j = \{(\mu_i(x_j)), \sigma_i(x_j)^2) \mid \forall i < l \} \;\forall j \in D_{train} $. Effectively, this computes the channel-wise mean and variance of the feature maps at each BN layer in $\phi$, across the dimensions of each individual datapoint. Finally, we evaluate the distance between the characteristic vector of each datapoint and the centroid characteristic vectors of each class, and assign as real label the class that minimizes such distance: 

\begin{equation}
\label{eq:y_real}
  y^r_j = \argmin_y d(\mathcal{X}_j, \mathcal{C}_y) \; \forall x_j \in D_{train}
\end{equation}

where $d$ is a distance metric. Whenever $y^r_j \neq y_j$ for a given datapoint $x_j$, i.e., the real label differs from the dataset label, we consider $x_j$ a potential poison and remove it from the dataset. Therefore, the clean training set is defined as:

\begin{equation}
  D_{clean} = \{ (x_j, y_j) \mid y^r_j = y_j \forall j < N\}
\end{equation}

We show that our approach is not only effective in isolating a clean dataset $D_{clean}$, but also that the subset of poisons $P$ which are not detected by our algorithm are in fact \emph{failed poisons}: perturbed datapoints that, when trained on, do not poison the model.

\subsubsection{Distance Metric}
The distance metric $d$ in Eq.~\ref{eq:y_real} measures the distance between a datapoint and the centroid of each class at different depths in the network and aggregates them in a single value. It is defined as follows:

\begin{equation}
\label{eq:distance}
    d(\mathcal{X}_j, \mathcal{C}_y) = \sum_{i=0}^l \gamma_i \: (\beta \: sim(\mu_i(x_j), \mu_i(X_y)) + (1-\beta) \: sim(\sigma_i(x_j)^2, \sigma_i(X_y)^2))
\end{equation}

where $\gamma_i$ is a coefficient defining the weight for each BN layer and $\beta$ defines the weight of the BN mean and variance in the computation. The function $sim$ in Eq.~\ref{eq:distance} can be any appropriate similarity metric between vectors. We used cosine distance: 

\begin{equation}
    sim(A,B) = 1 - \frac{A \cdot B}{\|A\|\|B\|}
\end{equation}

\section{Experimental Setup}
\label{sec:setup}
This section describes the experimental setup and dataset used to evaluate our proposed approach, as well as the state of the art approaches we compare against.

\subsection{Dataset}
\label{sec:dataset}
We use two image dataset in our experimental evaluation: CIFAR10~\cite{krizhevsky2009learning} and CINIC10~\cite{darlow2018cinic}. CIFAR10 consists of 60,000 color images of 32x32 pixel dimensions equally divided in 10 classes, split 50,000 for the training set and 10,000 for the testing set. The CINIC10 dataset is a superset of CIFAR10 that includes images from the ImageNet dataset~\cite{deng2009imagenet} downsampled to the same 32x32 pixel dimensions as the original CIFAR10 images. CINIC10 has a total of 270,000 color images equally split in the same 10 classes of CIFAR10. CINIC10 is split in three equal-sized subsets of 90,000 images: training, validation and testing. CINIC10 is designed as a drop-in replacement for CIFAR10 to train on the same task and has a similar but different distribution~\cite{darlow2018cinic}, making it a good candidate for a transfer learning setting.

\subsection{Poison Generation Algorithms and Defenses}
\label{sec:sota}
Similar to related works in the area of triggerless clean-label attacks~\cite{geiping2021strong,yang2022not}, we use the following poisoning algorithms in our evaluation: Feature Collison (FC)~\cite{shafahi2018poison}, Bullseye Polytope (BP)~\cite{aghakhani2021bullseye}, and Gradient Matching (GM)~\cite{geiping2020witches}, which we describe in Section~\ref{sec:background}. When possible, we use the original implementation from the authors, otherwise, we use the implementation by Schwarzschild et al.~\cite{schwarzschild2021just}.

We compare against three existing poison detection approaches:
Spectral Signatures~\cite{tran2018spectral}, Deep-KNN~\cite{peri2020deep}, and EPIC~\cite{yang2022not}. Spectral Signatures, proposed by Tran et al., is a seminal work in the area and is often used for comparison. Deep-KNN by Peri et al. is based on feature-space clustering, and is often used as a comparison point in the transfer learning setting. 
Finally, EPIC by Yang et al. is the current state-of-the-art in clean-label poisoning detection. EPIC is a filtering technique that uses the gradient-space representation of the datapoints during training to detect and remove isolated points from the training set. We use the implementation of the defenses provided by the authors for KNN and EPIC, while for Spectral Signatures we use a more recent implementation by Fowl et al.~\cite{poisoning_github}.

\section{Evaluation}
This section presents the experimental evaluation of our poison detection and filtering approach. Under multiple experimental settings, we show that our proposed technique consistently outperforms other approaches in poison detection performance and final model accuracy. This section is structured as follows. In Section~\ref{sec:eval_distance} we analyze the distribution of clean and poisoned datapoints and show that our characteristic vector representation is effective in isolating malicious points. Section~\ref{sec:eval_detection} evaluates the poison detection performance of our approach compared to state-of-the-art under different experimental settings.

\subsection{Poisons vs Clean Samples: A Characteristic Vector Perspective}
\label{sec:eval_distance}
In this section, we analyze the distribution of poisoned datapoints generated by different algorithms through the lense of their characteristic vector. We show that poisons and clean points are easily separable in the characteristic vector space, and that poisons tend to reside in the same class manifold as the target class. We also show that poisoned characteristic vectors (i.e., characteristic vectors of poisoned datapoints) that overlap with the distribution of clean characteristic vectors in fact belong to failed poisons: poisoned datapoints that, when trained on, fail to poison the model. For all experiments in this section, we follow the experimental setup used in previous works~\cite{zhu2019transferable,aghakhani2021bullseye}. 
We use a ResNet18 feature extractor pre-trained on the CIFAR10 dataset using the first 4,800 images of each class. The poisons are generated using ``ship'' as the base class and ``frog'' as the target class using base images that are not part of the training set. The clean datapoints used for the plots are not part of the training set. 

\begin{figure*}[t]
   \centering
   \subfloat[Feature Collision All] {
     \includegraphics[width=.30\textwidth]{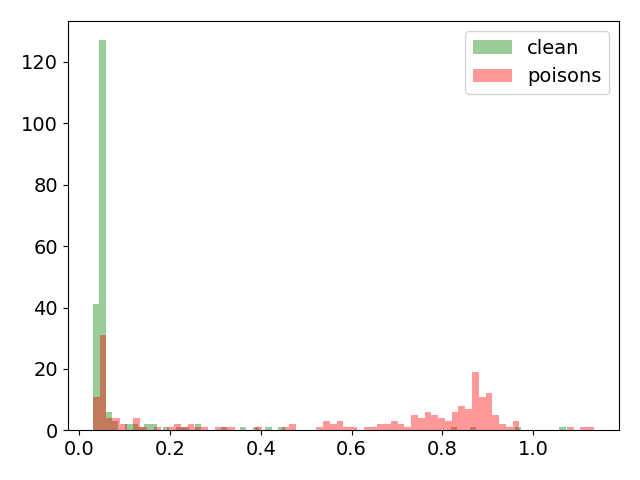}
     \label{fig:fc_all}
   }
   \subfloat[Bullseye Polytope All]{
     \includegraphics[width=.30\textwidth]{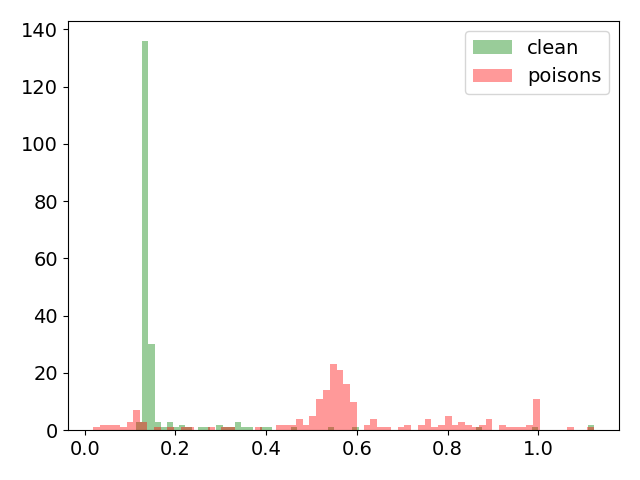}
     \label{fig:bp_all}
   }
   \subfloat[Gradient Matching All]{
     \includegraphics[width=.30\textwidth]{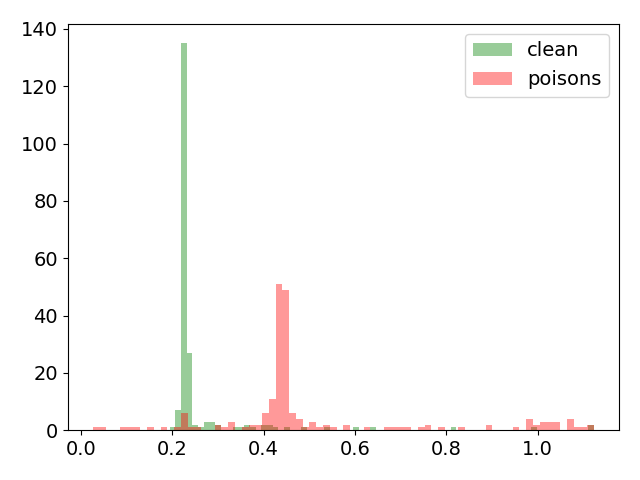}
     \label{fig:gm_all}
   }
   
   \medskip
   
   \subfloat[Feature Collision Real] {
     \includegraphics[width=.30\textwidth]{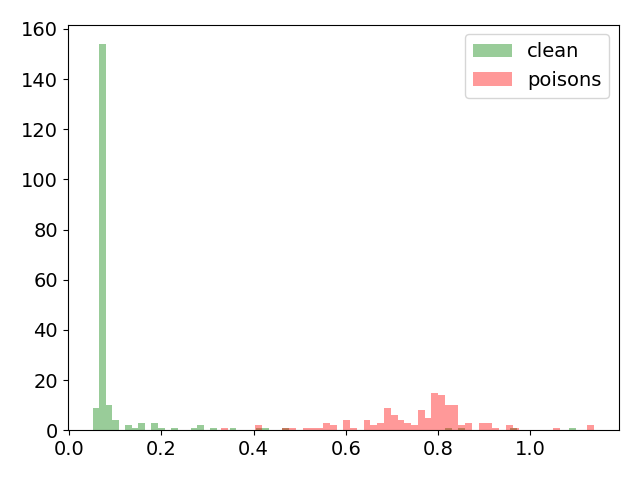}
     \label{fig:fc_real}
   }
   \subfloat[Bullseye Polytope Real]{
     \includegraphics[width=.30\textwidth]{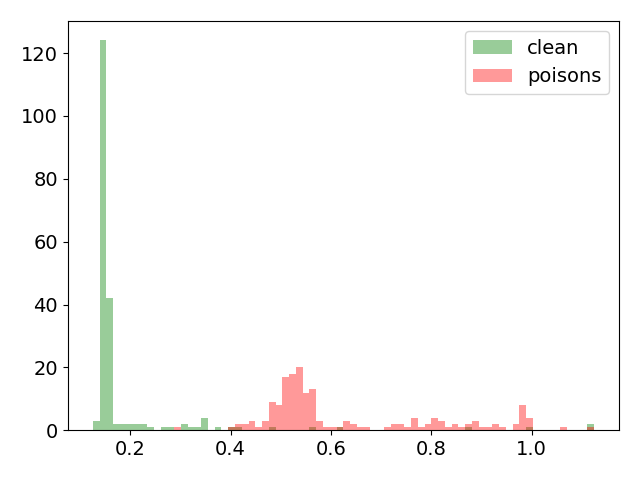}
     \label{fig:bp_real}
   }
   \subfloat[Gradient Matching Real]{
     \includegraphics[width=.30\textwidth]{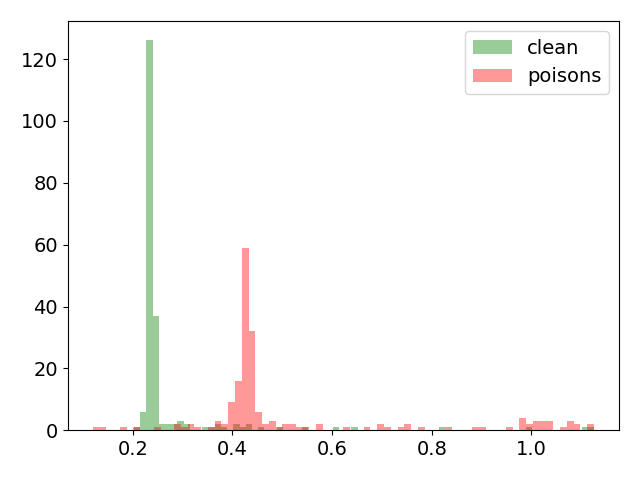}
     \label{fig:gm_real}
   }
   
   \caption{Computed distance $d$ (see Eq.~\ref{eq:distance}) between the characteristic vectors of poisoned and clean datapoints of the base class from the base class centroid. Figures~\ref{fig:fc_all},~\ref{fig:bp_all}, and~\ref{fig:gm_all} in the top row show the distance to the centroid for all poisons generated with FC, BP, and GM respectively. Figures~\ref{fig:fc_real},~\ref{fig:bp_real}, and~\ref{fig:gm_real} in the bottom row show the distance to the centroid only for real (i.e., effective) poisons.}
   \label{fig:distance}
   
   \vspace{-1.5em}
 \end{figure*}

Figure~\ref{fig:distance} plots the distribution of the distance from the base class centroid of 200 poisoned characteristic vectors, and 200 clean characteristic vectors beloging to the base class.
In the top row, Figures~\ref{fig:fc_all},~\ref{fig:bp_all}, and~\ref{fig:gm_all} show the distance for the characteristic vectors of all 200 generated poisons, while in the bottom row Figures~\ref{fig:fc_real},~\ref{fig:bp_real}, and~\ref{fig:gm_real} plot the distance only for real (i.e., effective) poisons. As depicted in the figure, we can see that the distance distribution of clean and poisoned characteristic vectors are easily separated and the overlap is minimal. This demonstrates that characteristic vectors effectively capture the shift in feature-level distribution caused by different poisoning attacks. Moreover, if we compare the top and bottom rows of Figure~\ref{fig:distance}, we can see that the overlapping characteristic vectors belong to failed poisons: perturbed base images that, when trained on, do not poison the neural network, nor degrade its performance. 
This further suggests that characteristic vectors describe intrinsic properties of the distribution of datapoints of a given class.

\begin{figure*}[t]
   \centering
   \subfloat[Feature Collision\label{fig:fc_proj}] {
     \includegraphics[width=.49\textwidth]{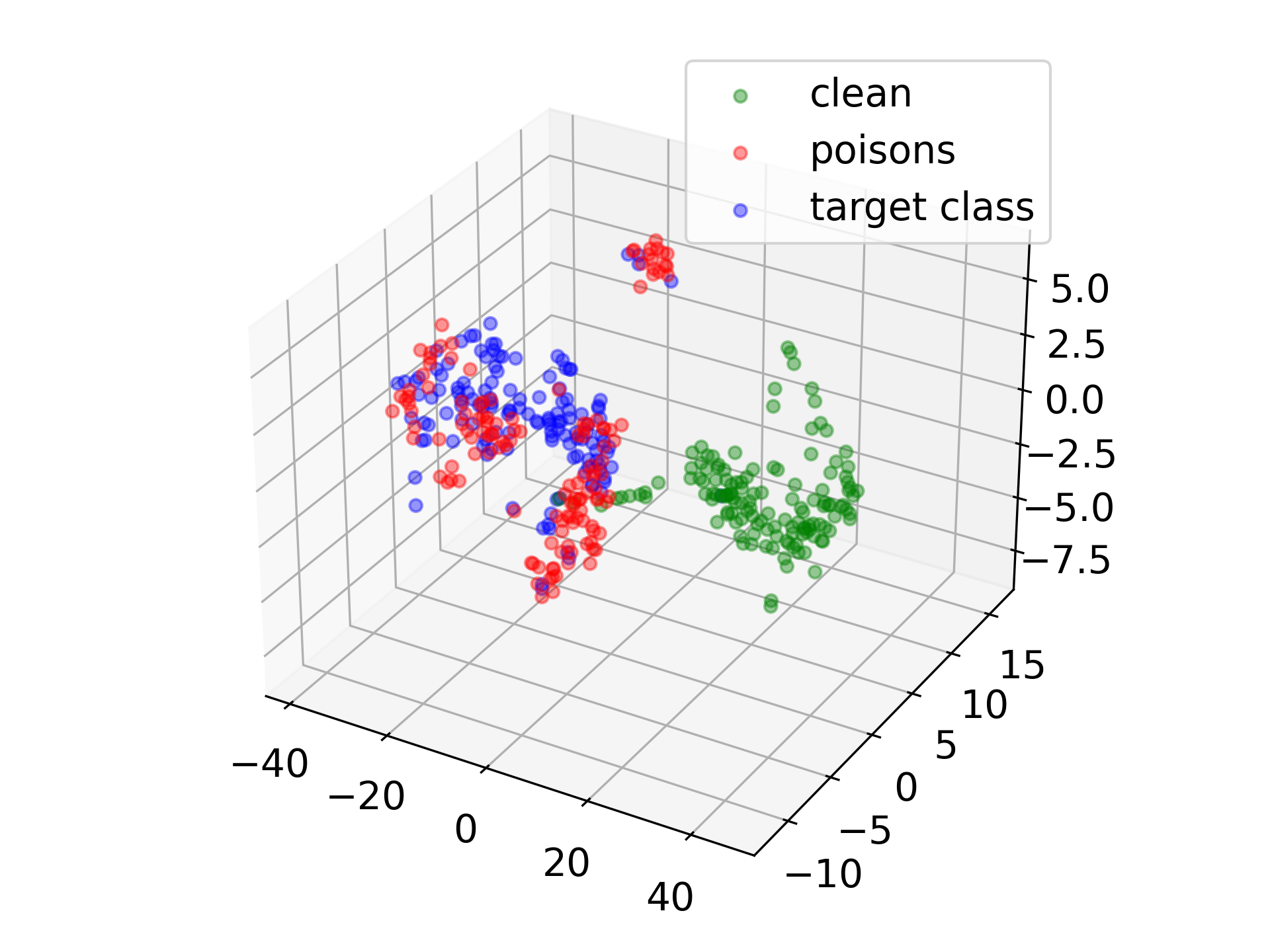}
   }
   \subfloat[Bullseye Polytope\label{fig:bp_proj}]{
     \includegraphics[width=.49\textwidth]{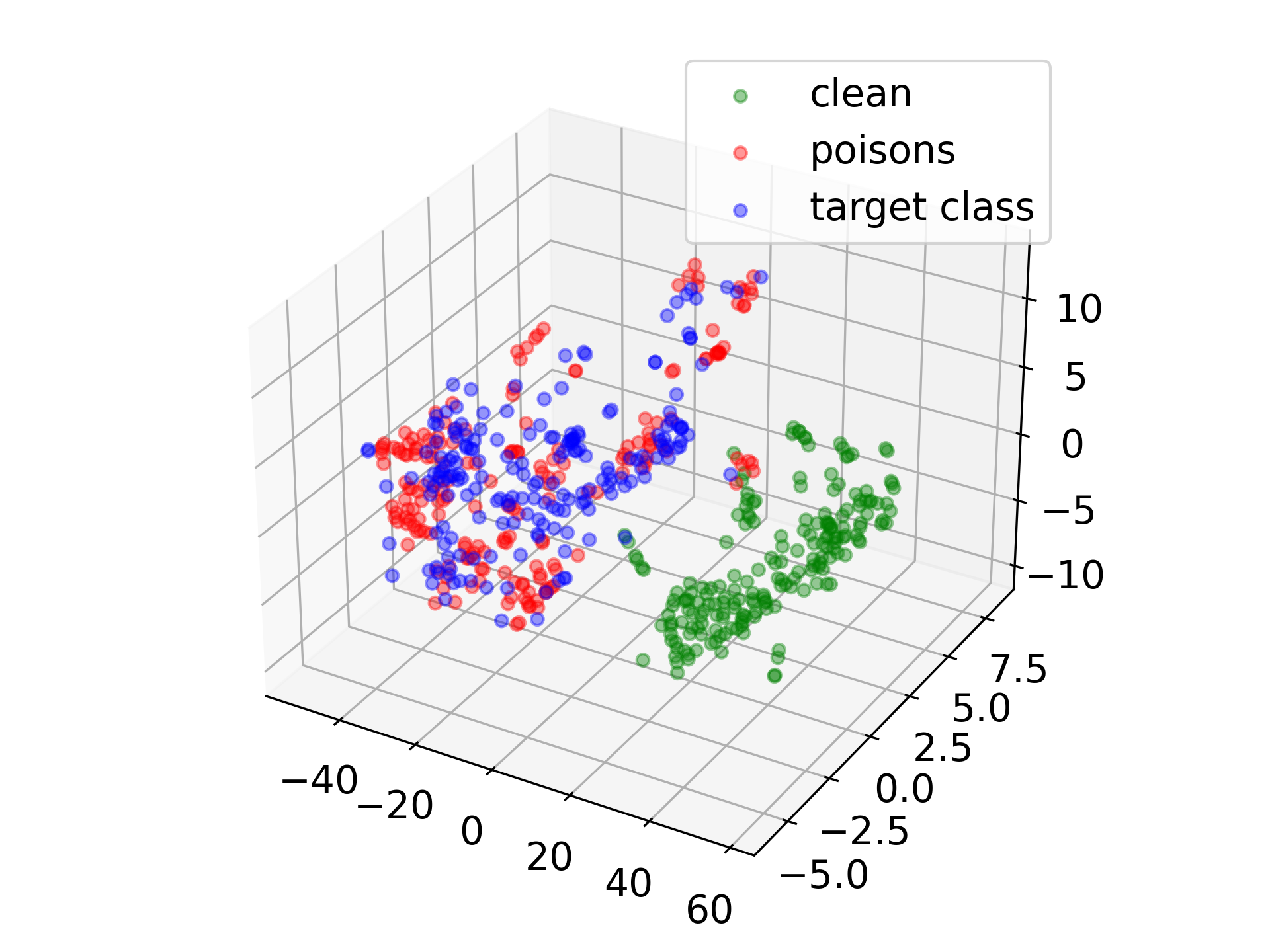}
   }
   
   \medskip
   
   \subfloat[Gradient Matching\label{fig:gm_proj}]{
     \includegraphics[width=.49\textwidth]{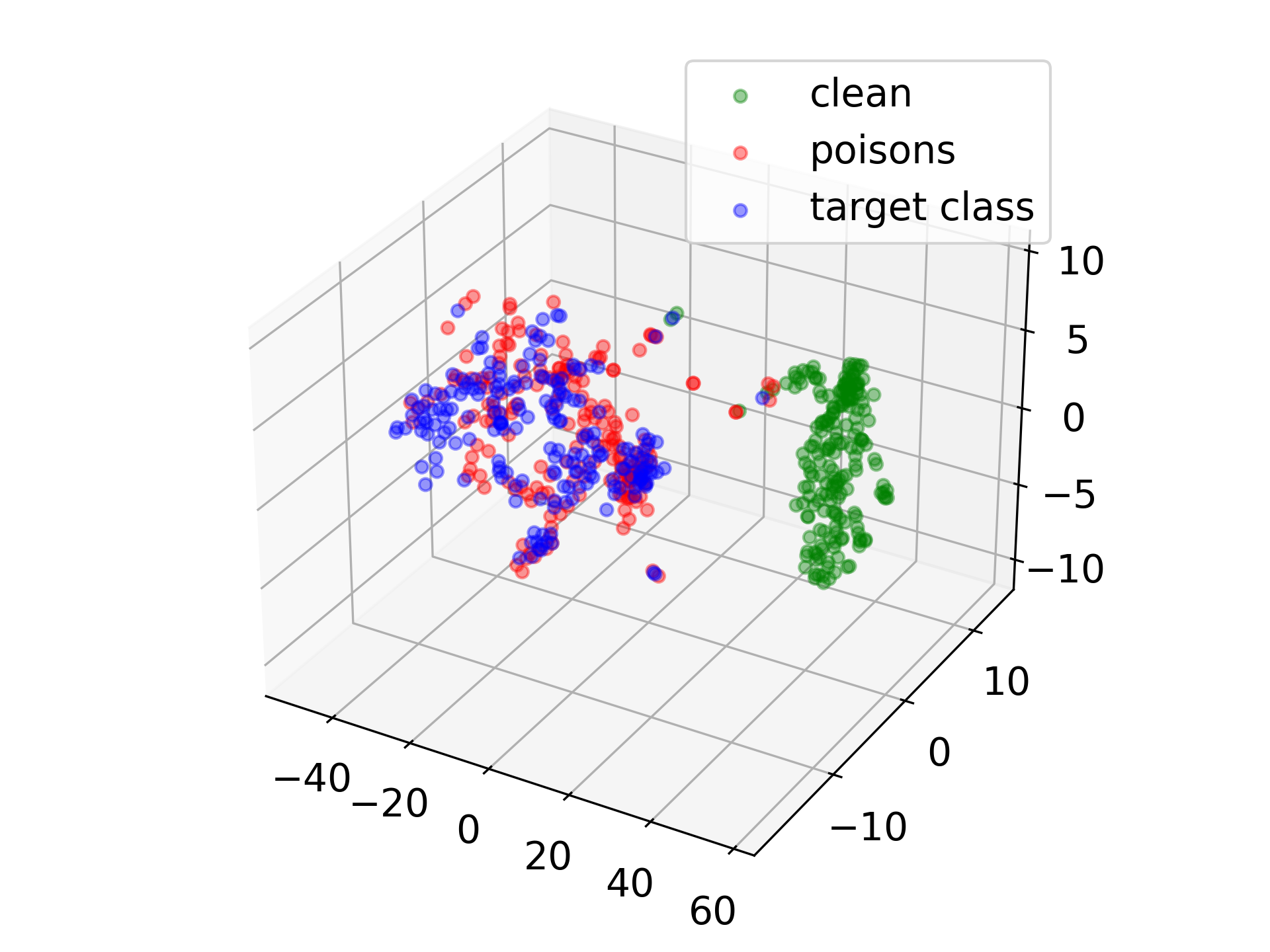}
   }
   
   \caption{Projection of the characteristic vector of base class clean datapoints, base class poisoned datapoints, and target class datapoints. Computed only for real poison samples.}
   \label{fig:projection}
 \end{figure*}

We further validate our hypothesis that poisoned datapoints are disjointed from the distribution of the base class and reside in the class manifold of the target class in the characteristic vector space. Figure~\ref{fig:projection} illustrates the projection of poisoned datapoints, clean datapoints belonging to the base class, and clean datapoints  belonging to the target class in the characteristic vector space. As we can see, for all considered poisoning algorithms, the clean datapoints are clearly separated from the poisoned datapoints. Furthermore, the distribution of poisoned datapoints overlaps almost exactly with the distribution of the target class datapoints in the characteristic vector space. This result validates our hypothesis and explains the effectiveness of our distance-based poison detection approach. To generate effective poisons, poisoning algorithms create perturbations that push the base images away from the base class and toward the target class. While certain poison generation algorithms such as FC explicitly promote this objective, our analysis shows that in the characteristic vector space, this behavior generalizes to other approaches as well. Finally, Figures~\ref{fig:distance} and~\ref{fig:projection} highlight why poisoning algorithms like GM are more effective than others, such as FC. By comparing Figures~\ref{fig:fc_all} and~\ref{fig:fc_real} we can see that a considerable portion of FC poisons are failed poisons, while for GM almost all generated poisons are real (Figures~\ref{fig:gm_all} and~\ref{fig:gm_real}). Moreover, we can see in Figure~\ref{fig:gm_proj} that GM poisons tend to be clustered and overlap almost exactly with the target class datapoints, while FC poisons are more spread out (Figure~\ref{fig:fc_proj}) and coalesce in sub-clusters that can be far from the target class datapoints.

\begin{table*}[t]
\scriptsize
\def\arraystretch{1.3}
\caption{Average success rate of FC, BP, and GM poison generation algorithms against multiple defenses and test accuracy for each defense in the CIFAR10 transfer learning setting. Poison budget: $14\%$ of dataset. Perturbation budgets range from 10/255 to 30/255. Lower success rate, higher test accuracy is better.}
\centering
\begin{tabular}[c]{ l | l | r | r | r | r | r | r | r | r | r }
\toprule
\multirow{3}{*}{\textbf{Attack}} & \multirow{3}{*}{\textbf{Architecture}} & \multicolumn{9}{c}{\textbf{Defense}} \\
\cline{3-11}
& & \multicolumn{3}{c|}{\textbf{KNN}~\cite{peri2020deep}} & \multicolumn{3}{c|}{\textbf{Spectral}~\cite{tran2018spectral}} & \multicolumn{3}{c}{\textbf{Ours}} \\
\cline{3-11}
& & \makecell[c]{Attack\\Succ.} & \makecell[c]{Test\\Acc.} & \makecell[c]{Clean\\Acc.} & \makecell[c]{Attack\\Succ.} & \makecell[c]{Test\\Acc.} & \makecell[c]{Clean\\Acc.} & \makecell[c]{Attack\\Succ.} & \makecell[c]{Test\\Acc.} & \makecell[c]{Clean\\Acc.} \\
\hline
\multirow{3}{*}{FC} 
& ResNet18 & 15.99 & 89.03 & 89.45 & 3.57 & 84.01 & 89.45 & 1.19 & 89.37 & 89.45  \\ \cline{2-11}
& ResNet50 & 11.42 & 89.14 & 89.50 & 5.47 & 80.06 & 89.50 & 4.28 & 89.41 & 89.50  \\ \cline{2-11}
& MobilenetV2 & 6.14 & 90.31 & 90.22 & 2.43 & 87.80 & 90.22 & 6.14 & 90.21 & 90.22  \\ \cline{2-11}
& Densenet121 & 0.00 & 89.39 & 89.38 & 0.00 & 88.58 & 89.38 & 0.00 & 89.35 & 89.38  \\ \cline{2-11}
& \textbf{Average} & \textbf{8.39} & \textbf{89.47} & \textbf{89.64} & \textbf{2.87} & \textbf{85.11} & \textbf{89.64} & \textbf{2.90} & \textbf{89.59} &  \textbf{89.64} \\ 

\midrule

\multirow{3}{*}{BP} 
& ResNet18 & 95.56 & 87.06 & 89.45 & 74.44 & 65.14 & 89.45 & 5.56 & 89.38 & 89.45  \\ \cline{2-11}
& ResNet50 & 98.89 & 86.59 & 89.50 & 90.00 & 73.90 & 89.50 & 6.67 & 89.38 & 89.50  \\ \cline{2-11}
& MobilenetV2 & 30.00 & 87.52 & 90.22 & 7.78 & 60.98 & 90.22 & 4.44 & 90.18 & 90.22  \\ \cline{2-11}
& Densenet121 & 39.75 & 88.54 & 89.38 & 49.26 & 67.93 & 89.38 & 0.00 & 89.33 & 89.38  \\ \cline{2-11}
& \textbf{Average} & \textbf{66.05}	& \textbf{87.43} & \textbf{89.64} & \textbf{55.37} & \textbf{66.99} & \textbf{89.64} & \textbf{4.17} & \textbf{89.57} & \textbf{89.64} \\ 

\midrule

\multirow{3}{*}{GM} 
& ResNet18 & 48.15 & 87.57 & 89.45 & 74.32 & 64.73 & 89.45 & 3.33 & 89.39 & 89.45  \\ \cline{2-11}
& ResNet50 & 48.81 & 86.96 & 89.50 & 84.02 & 70.80 & 89.50 & 6.82 & 89.37 & 89.50  \\ \cline{2-11}
& MobilenetV2 & 33.10 & 86.96 & 90.22 & 61.11 & 60.97 & 90.22 & 3.41 & 90.09 & 90.22  \\ \cline{2-11}
& Densenet121 & 60.86 & 88.26 & 89.38 & 44.69 & 61.82 & 89.38 & 2.22 & 89.35 & 89.38  \\ \cline{2-11}
& \textbf{Average} & \textbf{47.73} & \textbf{87.44} & \textbf{89.64} & \textbf{66.04} & \textbf{64.58} & \textbf{89.64} & \textbf{3.95} & \textbf{89.55} & \textbf{89.64} \\ 

\bottomrule
\end{tabular}
\label{tab:avg_all_cifar}
\end{table*}

\subsection{Poison Detection}
\label{sec:eval_detection}
This section evaluates the effectiveness of our approach in preventing model poisoning and preserving test accuracy. We compare against several existing approaches and show that our technique outperforms them under multiple experimental conditions. We consider two different transfer learning settings: transfer learning on different subsets of CIFAR10 as considered in previous works~\cite{shafahi2018poison,zhu2019transferable,aghakhani2021bullseye}, and CINIC10 to CIFAR10 transfer learning. In the following sections, we present the experimental setup in detail and discuss our results.

\vspace{-1em}

\subsubsection{CIFAR10 Transfer Learning}
This section presents our results in the CIFAR10 transfer learning setting. We use the same experimental setup as related works~\cite{zhu2019transferable,aghakhani2021bullseye,peri2020deep}. We pre-train the feature extractor model $\phi$ on CIFAR10 using the first 4,800 images of each class. Of the remaining images, the first 50 for each class are used as the fine-tuning dataset for transfer learning ($D_{train}$ in Section~\ref{sec:formal_description}). The base class used to create poisons is ``ship'' and the target class is ``frog''. Results are averaged over 30 different target samples which are not part of the training nor fine-tuning sets (indices 4950 to 4980). We leave the test set unchanged to allow direct comparisons of test accuracy. During transfer learning the feature extractor is frozen and only the model $f$ is trained (see Section~\ref{sec:threat_model}). The fine-tuning is done using the Adam optimizer with a learning rate of 0.1 for 60 epochs. 

Table~\ref{tab:avg_all_cifar} shows the results of our evaluation. We test our proposal and existing approaches against FC, BP, and GM poisons across different feature extractor architectures and perturbation budgets between $10/255$ and $30/255$. The performance for all defenses is reported only on poisons that lead to successful attacks (i.e., the undefended attack success rate is $100\%$). The test accuracy indicates the accuracy on the CIFAR10 test set of the model $f$ trained on the fine-tuning dataset filtered with a given defense. The clean accuracy is the accuracy on the CIFAR10 test set of the model $f$ trained only on clean data from the fine-tuning set. As we can see, on average our proposed approach outperforms existing defenses both in poison detection performance and test accuracy. Across all architectures and poisoning algorithms, our technique significantly reduces attack success rate to an average of $3.67\%$ (vs $100\%$ undefended), with negligible loss in test accuracy. Existing approaches fare well against weaker attacks such as FC, but consistently fail to defend the model against BP and GM, with attack success rates reaching up to $\sim60\%$. Moreover, Spectral in particular considerably degrades test accuracy when BP and GM poisons are used. On the contrary, our approach effectively filters poisoned datapoints even against stronger attacks, with an average attack success rate of $4.17\%$ and $3.95\%$ for BP and GM respectively, and no impact on testing performance. Due to space limitations, we include additional detailed results and plots in Appendix~\ref{app:results}.

\begin{table*}[t]
\scriptsize
\def\arraystretch{1.3}
\caption{Average success rate of FC, BP, and GM attacks against our approach and EPIC. ResNet18 architecture in the CIFAR10 transfer learning setting. Poison budget: $14\%$ of dataset. Perturbation budgets range from 10/255 to 30/255. Lower success rate, higher test accuracy is better.}
\centering
\begin{tabular}[c]{ l | r | r | r | r | r | r | r | r | r }
\toprule
\multirow{3}{*}{\textbf{Defense}} & \multicolumn{9}{c}{\textbf{Attack}} \\
\cline{2-10}
& \multicolumn{3}{c|}{\textbf{FC}} & \multicolumn{3}{c|}{\textbf{BP}} & \multicolumn{3}{c}{\textbf{GM}} \\
\cline{2-10}
& \makecell[c]{Attack\\Succ.} & \makecell[c]{Test\\Acc.} & \makecell[c]{Clean\\Acc.} & \makecell[c]{Attack\\Succ.} & \makecell[c]{Test\\Acc.} & \makecell[c]{Clean\\Acc.} & \makecell[c]{Attack\\Succ.} & \makecell[c]{Test\\Acc.} & \makecell[c]{Clean\\Acc.} \\
\hline
EPIC(0.1) Adam & 0.00 & 70.27 & 89.45 & 0.00 & 72.08 & 89.45 & 0.00 & 70.66 & 89.45  \\ \hline
EPIC(0.2) Adam & 0.00 & 69.77 & 89.45 & 0.00 & 69.73 & 89.45 & 0.00 & 70.20 & 89.45  \\ \hline
EPIC(0.3) Adam & 0.00 & 28.18 & 89.45 & 0.00 & 24.81 & 89.45 & 0.00 & 32.78 & 89.45  \\ \hline
EPIC(0.1) SGD & 0.00 & 72.69 & 89.45 & 0.00 & 73.90 & 89.45 & 0.00 & 73.04 & 89.45  \\ \hline
EPIC(0.2)  SGD & 0.00 & 71.21 & 89.45 & 0.00 & 72.37 & 89.45 & 0.00 & 71.98 & 89.45  \\ \hline
EPIC(0.3) SGD & 0.00 & 15.71 & 89.45 & 0.00 & 16.38 & 89.45 & 0.00 & 15.98 & 89.45  \\ \hline \midrule
\textbf{Ours} & \textbf{1.19} & \textbf{89.37} & \textbf{89.45} & \textbf{5.56} & \textbf{89.38} & \textbf{89.45} & \textbf{3.33} & \textbf{89.39} & \textbf{89.45}  \\
\bottomrule
\end{tabular}
\label{tab:avg_epic}
\end{table*}

\textbf{EPIC.} Table~\ref{tab:avg_epic} compares our approach to EPIC in the CIFAR10 transfer learning setting under different conditions. As we can see, in all our tests EPIC reduces the average attack success rate to 0 for all considered attacks. While this result is remarkable, it is achieved at the expense of the final model's performance. We tested EPIC with different suggested values for the subset of medoids selected at each iteration~\cite{yang2022not}, shown between brackets in the table. We also tested the defense using the SGD optimizer for transfer learning as done in the original paper, rather than Adam. Under all considered scenarios, the test accuracy of the final model when using EPIC degrades considerably. On the other hand, our approach consistently maintains high test performance, while also greatly reducing poisoning success rate. We note that our results differ from those reported in the original EPIC paper~\cite{yang2022not}. This discrepancy is due to the different transfer learning settings adopted. In the original EPIC paper, an atypical transfer learning setting is used where the \emph{full} CIFAR10 trainset is used also as the fine-tuning set for the model $f$. In this paper, we use the same transfer learning setting proposed by previous works on poisoning attacks and defense~\cite{zhu2019transferable,aghakhani2021bullseye,peri2020deep}, where the final model $f$ is fine-tuned on a \emph{small, separate} set of points that are not in the train set of the feature extractor. 

\vspace{-1em}

\subsubsection{CINIC10 Transfer Learning}

\begin{figure*}[tb]
   \centering
   \subfloat[Feature Collision] {
     \includegraphics[width=.31\textwidth]{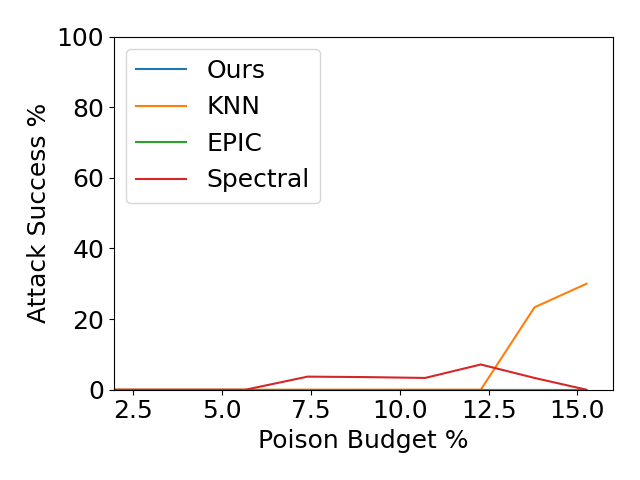}
     \label{fig:fc_graph}
   }
   \subfloat[Bullseye Polytope]{
     \includegraphics[width=.31\textwidth]{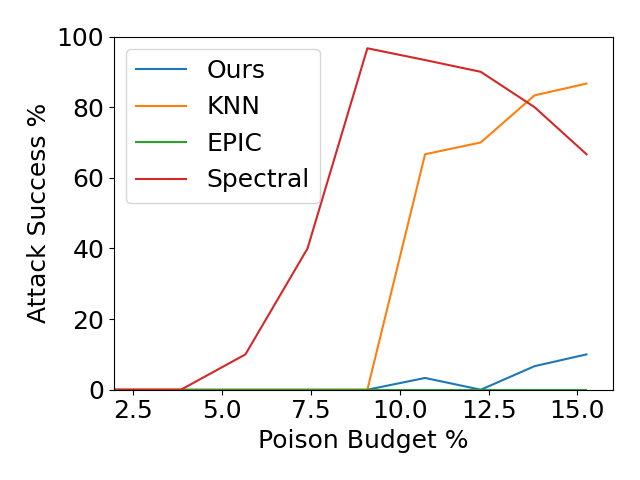}
     \label{fig:bp_graph}
   }
   \subfloat[Gradient Matching]{
     \includegraphics[width=.31\textwidth]{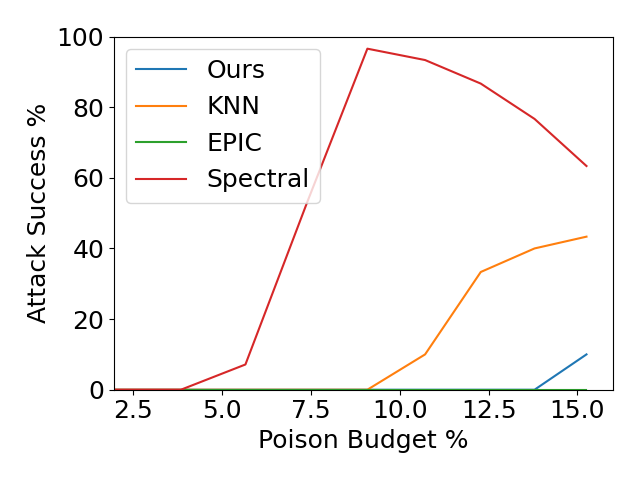}
     \label{fig:gm_graph}
   }
   
   \caption{Average poisoning success rate of FC, BP, and GM attacks on ResNet18 against multiple defenses in the CINIC10 transfer learning setting, with a perturbation budget of 20/255, for different poison budgets. Lower is better.}
   \label{fig:def_plots}
 \end{figure*}

Typically, when doing transfer learning the fine-tuning set is sampled from a (slightly) different distribution than the training set used for the feature extractor. In the previous section, we evaluated our approach in a setting that is consistent with previous art. However, such a setting is not representative of ``true'' transfer learning~\cite{schwarzschild2021just}, as the fine-tuning set has the same distribution as the training set used for the extractor. In this section, we evaluate our poison filtering technique in a transfer learning setting where the pre-train dataset has a different, but similar, distribution than the fine-tuning set. We pre-train the feature extractor $\phi$ on the training subset of CINIC10 and fine-tune the final model $f$ on a subset of the CIFAR10 dataset. Since CINIC10 is a superset of CIFAR10, we avoid overlaps by sampling the fine-tuning images of CIFAR10 from the validation subset of CINIC10, which is not used in the training of $\phi$. As in previous evaluations, we select 50 images from each class for fine-tuning and use the same base and target classes (``ship'' and ``frog'', respectively). All results are averaged over 30 different target samples that are not part of the training or fine-tuning sets, and the results are reported only for poisons that lead to successful attacks. The fine-tuning is done using the Adam optimizer with a learning rate of 0.1 for 60 epochs.

\begin{table*}[t]
\scriptsize
\def\arraystretch{1.3}
\caption{Average success rate of FC, BP, and GM attacks against multiple defenses, and test accuracy for each defense. ResNet18 architecture in the CINIC10 transfer learning setting. Poison budget: $14\%$ of dataset. Perturbation budgets range from 10/255 to 30/255. Lower success rate, higher test accuracy is better.}
\centering
\begin{tabular}[c]{ l | r | r | r | r | r | r | r | r | r }
\toprule
\multirow{3}{*}{\textbf{Defense}} & \multicolumn{9}{c}{\textbf{Attack}} \\
\cline{2-10}
& \multicolumn{3}{c|}{\textbf{FC}} & \multicolumn{3}{c|}{\textbf{BP}} & \multicolumn{3}{c}{\textbf{GM}} \\
\cline{2-10}
& \makecell[c]{Attack\\Succ.} & \makecell[c]{Test\\Acc.} & \makecell[c]{Clean\\Acc.} & \makecell[c]{Attack\\Succ.} & \makecell[c]{Test\\Acc.} & \makecell[c]{Clean\\Acc.} & \makecell[c]{Attack\\Succ.} & \makecell[c]{Test\\Acc.} & \makecell[c]{Clean\\Acc.} \\
\hline
KNN & 27.78 & 86.82 & 87.83 & 82.22 & 84.98 & 87.83 & 35.71 & 86.27 & 87.83  \\ \hline
Spectral & 2.22 & 74.60 & 87.83 & 75.56 & 60.29 & 87.83 & 76.40 & 64.15 & 87.83  \\ \hline
EPIC(0.1) SGD & 0.00 & 72.44 & 87.83 & 0.00 & 69.90 & 87.83 & 0.00 & 70.37 & 87.83  \\ \hline \midrule
\textbf{Ours} & 0.00 & 87.65 & 87.83 & 4.44 & 87.49 & 87.83 & 2.22 & 87.49 & 87.83  \\

\bottomrule
\end{tabular}
\label{tab:avg_all_cinic}
\end{table*}

Figure~\ref{fig:def_plots} summarizes the results of our evaluation. It plots the attack success rate for all considered defenses against varying poison budgets on a ResNet18 network. As we can see, our approach consistently prevents poisoning across varying poison budgets, for all considered attacks. The only defense with comparable results is EPIC, but as we will discuss shortly, such results are achieved at the expense of a major performance penalty on the testing set. KNN defense is generally effective at lower poison budgets, but quickly fails when the attacker is allowed more poisons. Finally, Spectral Signatures is fairly effective against FC poisons, but fails to successfully defend against stronger attacks. We note that for both BP and GM attacks, the attack success rate against Spectral Signatures begins to decrease starting at $\sim8\%$ poison budget. While this behavior seems counter-intuitive, it is explained by a similar trend in testing accuracy. Effectively, for higher poison budgets Spectral Signatures discards a larger percentage of the fine-tuning set, resulting in lower attack success rates but also in major performance penalty for the final model.
Table~\ref{tab:avg_all_cinic} reports detailed results of our evaluation for a poison budget of $14\%$. As we can see, the results are similar to those reported in Tables~\ref{tab:avg_all_cifar} and~\ref{tab:avg_epic}. KNN performs best against FC and consistently fails against BP and GM, with a test accuracy that is marginally lower than clean accuracy. Spectral Signatures continues to perform well against weak attacks such as FC, but consistently fails against stronger attacks. The test performance penalty also remains high across all experiments. Finally, EPIC successfully detects and filters all poisons, but heavily penalizes the final model's performance. Similar to previous experiments, our approach consistently outperforms other techniques, preventing poisoning and maintaining test performance essentially unchanged.

\section{Related Works}
Adversarial attacks on machine learning~\cite{zhang2020adversarial,hitaj2022maleficnet,de2022evading} and robust defenses against such attacks~\cite{miller2020adversarial,piskozub2021malphase} have become popular topics in recent years, especially in critical domains such as cybersecurity~\cite{de2022reliable,hitaj2024you,pagnotta2022passflow}.
In the area of model poisoning, defenses can be categorized into sanitization (filtering) defenses and robust training methods. Filtering defenses aim to detect and remove poisoned datapoints from the training set before training the model, while robust training methods employ several training techniques to obtain clean models even when trained with malicious data. 
Robust training methods use a variety of techniques to ensure model robustness, such as strong data augmentation~\cite{borgnia2021strong}, randomized smoothing~\cite{weber2023rab}, gradient shaping~\cite{hong2020effectiveness}, and adversarial training on poisons~\cite{geiping2021strong}. Other robust training proposals exploit ensemble models and dataset partitioning to prevent poisoning~\cite{levine2020deep}, or ad-hoc training approaches such as differentially private SGD~\cite{ma2019data} and gradient ascent to revert the effect of poisons~\cite{li2021anti}. 
Sanitization-based defenses use many different features to detect poisons and filter the training set. Tran et al~\cite{tran2018spectral} detect backdoor triggers based on their correlation with the top singular vector of the covariance matrix of learned representations. Other approaches isolate datapoints based on a radial distance in the feature space~\cite{steinhardt2017certified} and neuron activation patterns~\cite{chen2018detecting}, or based on feature space representation clustering~\cite{peri2020deep}. Finally, some techniques filter datapoints based on their projection in the gradient space during the training procedure, removing points that are isolated~\cite{yang2022not}.

Current defenses, both robust training and sanitization-based, have different shortcomings. Many defenses are designed against specific attacks and fail to generalize to different poison-generation approaches~\cite{paudice2019label,peri2020deep,chen2018detecting}. Other approaches are effective against some poisoning attacks, but fail when faced with stronger poison creation algorithms~\cite{koh2022stronger,shokri2020bypassing,tran2018spectral}. Finally, when applied in different settings, some proposals severely impact the trained model's performance~\cite{yang2022not,geiping2021strong}, or fail when adversaries have a large perturbation budget. In comparison, our proposed defense generalizes to different poison-generation approaches, is effective against strong attacks and large perturbation budgets, and does not affect the performance of the final model.

\section{Conclusions and Future Work}
We proposed a new defense against clean-label poisoning attacks in the transfer learning setting based on the idea of characteristic vectors. We proposed a new characteristic vector representation that effectively captures and describes key features of the datapoints, allowing us to differentiate poisons and clean samples in the characteristic vector space. We demonstrated that our representation allows us to differentiate real and failed poisons, and that real poisons reside in the data manifold of the target class in the characteristic vector space. Through extensive experimental evaluation, we demonstrated that our approach successfully detects and removes poisons from the training set without impacting the final model's performance. We compared against current state-of-the-art defenses in different experimental settings and showed that our approach outperforms them both in test accuracy and attack success rate.

As future work, we plan to extend our approach to the train-from-scratch scenario. Currently, our approach requires a pre-trained feature extractor to build characteristic vectors, and can therefore only be used in the transfer learning setting. We plan to study an iterative training approach to extend the applicability of our defense to all training settings.

\bibliographystyle{splncs04}
\bibliography{references}

\appendix
\begin{algorithm}[h]
\scriptsize
\DontPrintSemicolon
\KwInput{Model: $\phi$}
\KwOutput{Centroids: $\mathcal{C}$}
\KwData{Dataset: $D_{train}$, Classes: $Y$}
    $\mathcal{C} \leftarrow list(len(Y))$ \;
    \ForEach{$y \in Y$} {
        $X_y \leftarrow \{x_i \mid y_i == y \; \forall (x_i, y_i) \in D_{train} \}$ \;
        $\mathcal{C}_y \leftarrow list()$ \;
        \ForEach{$L_i^{bn} \in \phi$} {
            $\mathcal{C}_y \leftarrow append(\mathcal{C}_y, (\mu_i(X_y),\sigma_i(X_y)^2))$ \;
        }
    }
\caption{Centroid Computation}
\label{alg:centroids}
\end{algorithm}

\vspace{-4em}

\begin{algorithm}[h]
\scriptsize
\DontPrintSemicolon
\KwInput{Model: $\phi$, Centroids: $\mathcal{C}$}
\KwOutput{Dataset: $D_{Clean}$}
\KwData{Dataset: $D_{train}$, Classes: $Y$}

    $y^r \leftarrow zeroes(len(D_{train})$ \;
    $D_{clean} \leftarrow set()$ \;
    \ForEach{$x_i, y_i \in D_{train}$} {
        $\mathcal{X}_i \leftarrow list()$ \;
        \ForEach{$L_i^{bn} \in \phi$} {
            $\mathcal{X}_i \leftarrow append(\mathcal{X}_i, (\mu_i(x_i),\sigma_i(x_i)^2))$ \;
        }
        $dist \leftarrow inf(len(Y))$ \;
        \ForEach{$y \in Y$} {
            $dist[y] \leftarrow distance(\mathcal{C}_y, \mathcal{X}_i)$ \;
        }
        $y^r_i \leftarrow argmin(dist)$ \;
        \uIf{$y_i == y^r_i$} {
            $add(D_{clean}, (x_i, y_i))$ \;
        }
    }
\caption{Poison Filtering}
\label{alg:poison_filtering}
\end{algorithm}


\section{Implementation Details}
Algorithms~\ref{alg:centroids} and~\ref{alg:poison_filtering} show the pseudo-code for the centroid computation and poison filtering respectively. 
Algorithm~\ref{alg:centroids} takes as input the pre-trained feature extractor $\phi$. For each class $y$ in the dataset, it flows the full set of datapoints $X_y$ through the feature extractor and computes the characteristic vector of the centroid pseudo-datapoint $\mathcal{C}_y$ for that class using BN mean and variance at each layer.
Finally, the list of computed centroids $\mathcal{C}$ is given as output.
Algorithm~\ref{alg:poison_filtering} takes as input the pre-trained feature extractor $\phi$ and the previously computed centroid characteristic vectors $\mathcal{C}$ for all classes. For each datapoint in the training set, the characteristic vector $\mathcal{X}_i$ is computed by flowing each datapoint $x_i$ through the network and computing BN statistics at each BN layer in the network. Finally, the distance between the characteristic vector of each datapoint and the centroid characteristic vector of every class is computed following Eq.~\ref{eq:distance}, and the real label $y^r_i$ of the datapoint is defined as the label of the centroid with minimal distance to $x_i$ (Eq.~\ref{eq:y_real}). Lastly, the clean dataset $D_{clean}$ is populated with the set of datapoints for which the computed real label $y^r_i$ equals the dataset label $y_i$.

\section{Additional Experimental Results}
\begin{figure*}[h]
   \centering
   \subfloat[FC epsilon 10/255] {
     \includegraphics[width=.30\textwidth]{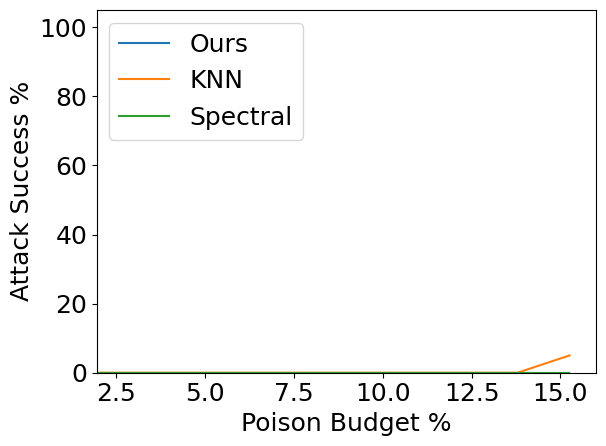}
     \label{fig:fc_all_10}
   }
   \subfloat[BP epsilon 10/255]{
     \includegraphics[width=.30\textwidth]{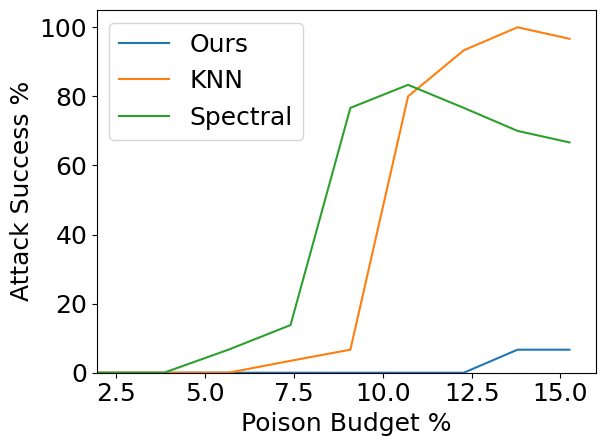}
     \label{fig:bp_all_10}
   }
   \subfloat[GM epsilon 10/255]{
     \includegraphics[width=.30\textwidth]{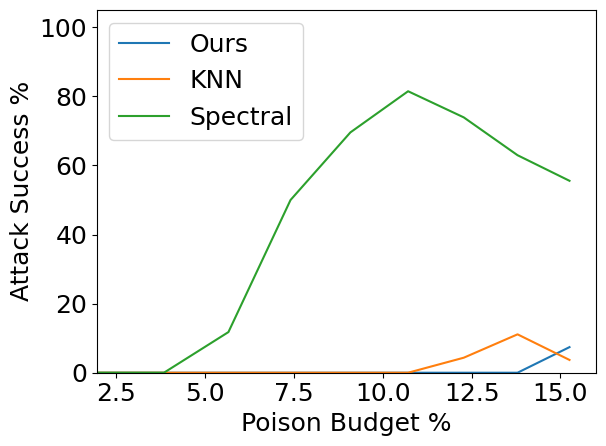}
     \label{fig:gm_all_10}
   }
   
   \smallskip
   
   \subfloat[FC epsilon 20/255] {
     \includegraphics[width=.30\textwidth]{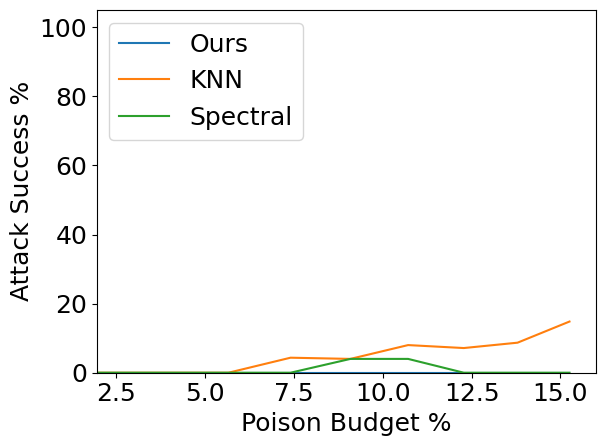}
     \label{fig:fc_all_20}
   }
   \subfloat[BP epsilon 20/255]{
     \includegraphics[width=.30\textwidth]{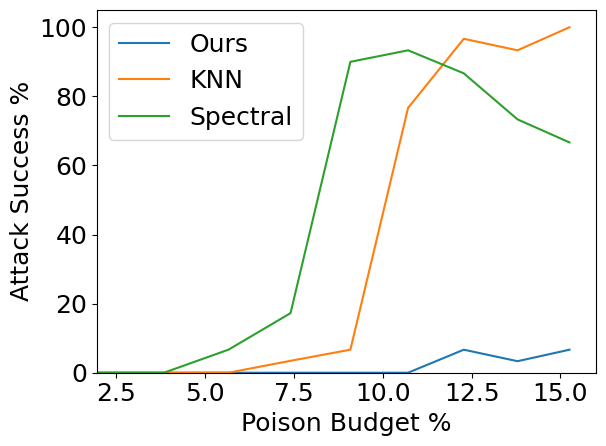}
     \label{fig:bp_all_20}
   }
   \subfloat[GM epsilon 20/255]{
     \includegraphics[width=.30\textwidth]{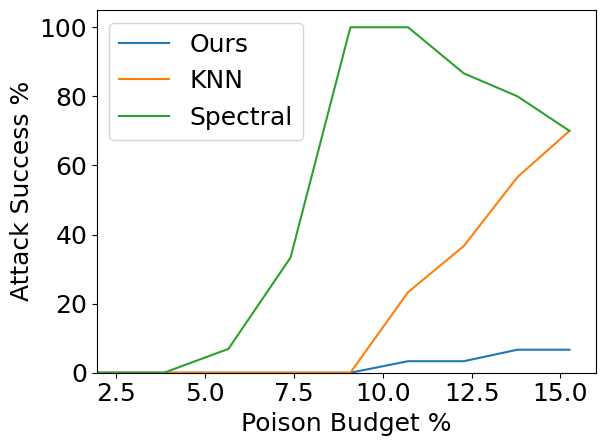}
     \label{fig:gm_all_20}
   }

   \smallskip
   
   \subfloat[FC epsilon 30/255] {
     \includegraphics[width=.30\textwidth]{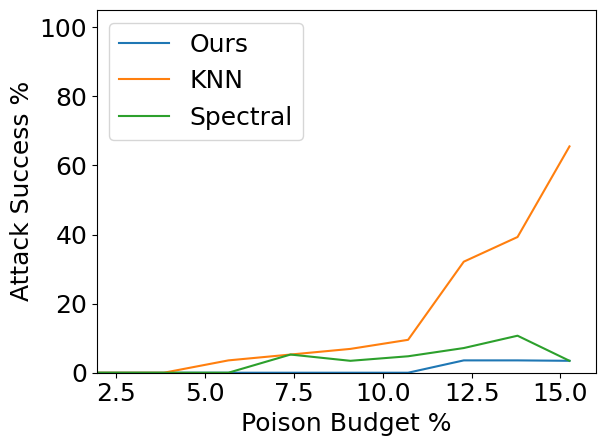}
     \label{fig:fc_all_30}
   }
   \subfloat[BP epsilon 30/255]{
     \includegraphics[width=.30\textwidth]{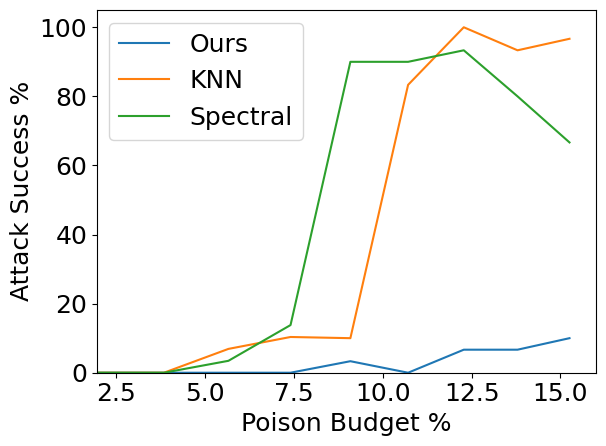}
     \label{fig:bp_all_30}
   }
   \subfloat[GM epsilon 30/255]{
     \includegraphics[width=.30\textwidth]{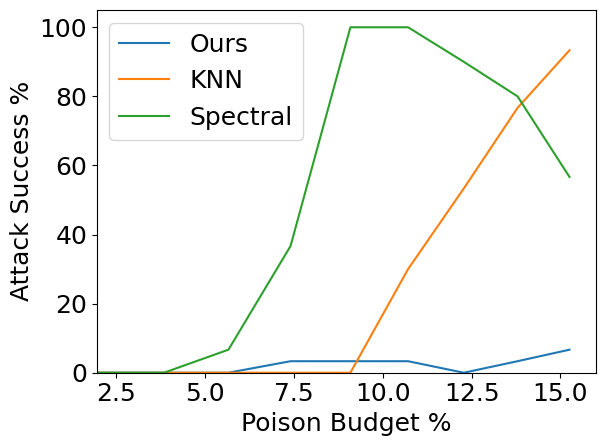}
     \label{fig:gm_all_30}
   }   
   
   \caption{Average poisoning success on ResNet18 against multiple defenses in the CIFAR10 transfer learning setting. Lower is better.}
   \label{fig:results_all}
 \end{figure*}

\vspace{-0.5em}
\label{app:results}
Figure~\ref{fig:results_all} illustrates the attack success rate against all defenses across different poison and perturbation budgets. EPIC defense is omitted, as the attack success rate against it is always $0\%$, with a large test accuracy penalty. See Section~\ref{sec:eval_detection} for a discussion of all results.

\vspace{-1.5em}

\end{document}